\documentclass[conference]{IEEEtran}
\usepackage{graphicx}
\usepackage{url}
\usepackage{amsmath}
\usepackage{amssymb}
\usepackage{tcolorbox}
\usepackage[utf8]{inputenc}
\usepackage[T1]{fontenc}
\usepackage[T2A]{fontenc}
\usepackage[russian,english]{babel}
\usepackage{enumitem}

\usepackage{graphicx}
\usepackage{float} 
\graphicspath{{results/}} 

\usepackage{amsmath}
\usepackage{amssymb}

\begin{document}

\title{ Metamorphic Testing for Evaluating Robustness of Sentiment Analysis Models}

\author{Sowjanya Karanam, Jayanth Bhargav}
\maketitle

\begin{abstract}
Machine learning systems are challenging to test using traditional software testing methods because their behavior is learned from data rather than explicitly programmed. This challenge is particularly significant in natural language processing (NLP), where small input variations can lead to unexpected model behavior. Conventional evaluation metrics, such as accuracy and F1-score, often fail to capture such inconsistencies. Metamorphic testing addresses this limitation by evaluating expected relationships between original and transformed inputs. With the increasing prevalence of generative artificial intelligence (GenAI) and language-based AI systems, the need for systematic robustness testing has become even more pronounced. Motivated by this growing need, this paper presents a lightweight, batch-based metamorphic testing approach for evaluating the robustness of such systems. In particular, we consider sentiment analysis as a representative NLP task and evaluate a pretrained classifier under four transformations: synonym replacement, typo injection, word-order variation, and negation. Using consistency score, flip accuracy, and confidence change, we identify strengths in handling semantic variations and vulnerabilities to textual noise and logical transformations. These findings highlight the value of metamorphic testing in identifying behavioral inconsistencies beyond conventional performance metrics and demonstrate its potential as a systematic framework for evaluating the robustness of language-based AI systems and guiding performance improvements before deployment.
\end{abstract}

\section{Introduction}

Machine learning systems are increasingly deployed in real-world applications such as recommendation systems, fraud detection, autonomous systems, and natural language processing. Unlike traditional software systems, which follow explicitly defined rules, machine learning models learn their behavior from data. As a result, their decision-making process is often complex, non-linear, and difficult to interpret. This creates significant challenges for testing and validation, as correct outputs are not always known in advance and model behavior may vary across different input distributions.

Traditional software testing techniques rely on the presence of a test oracle, which defines the expected output for a given input. However, in machine learning systems, especially in natural language processing tasks, defining such an oracle is often infeasible \cite{chen1998}. For example, determining the exact sentiment of a sentence can be subjective, and small variations in wording may not have a clearly defined correct output. As a result, machine learning systems are typically evaluated using statistical metrics such as accuracy, precision, recall, and cross-validation.

While these metrics provide a high-level measure of model performance, they are limited in their ability to capture how models behave under input variations \cite{zhang2019}. In real-world scenarios, input data is rarely clean or static. Text inputs may include paraphrasing, typographical errors, grammatical variations, or changes in word order. A model that performs well on a benchmark dataset may still produce inconsistent or unreliable predictions when exposed to such variations. This highlights the need for testing methods that evaluate not only predictive performance but also robustness and stability.

Metamorphic testing provides a practical approach to address the oracle problem in machine learning systems. Instead of requiring exact expected outputs, metamorphic testing defines relationships between inputs and outputs that should hold under specific transformations \cite{segura2016}. For example, if a sentence is modified by replacing a word with a synonym, the sentiment should typically remain unchanged. By verifying such relationships, metamorphic testing enables systematic evaluation of model behavior without requiring a complete oracle.

Recent work has demonstrated the effectiveness of metamorphic testing for validating machine learning and deep learning systems. Ding et al. \cite{ding2017} show how metamorphic testing can be used to validate deep learning frameworks, while the METTLE framework \cite{xie2020} extends this idea to assess machine learning systems under data transformations. However, much of the existing work focuses on structured data or theoretical validation, with limited emphasis on practical applications in natural language processing.

Natural language processing presents unique challenges for robustness testing due to the inherent variability of language. Sentiment analysis, in particular, is sensitive to linguistic nuances such as negation, emphasis, and context. A robust sentiment model should maintain consistent predictions under meaning-preserving transformations while correctly adapting to transformations that alter the underlying sentiment. Evaluating these properties requires testing approaches that go beyond traditional metrics and incorporate realistic input variations.

In this paper, we propose a lightweight, batch-based metamorphic testing framework for evaluating the robustness of sentiment analysis models. The approach applies a set of simple but effective linguistic transformations, including synonym replacement, typo injection, word-order variation, and negation. These transformations simulate common real-world variations in text data.

The proposed method evaluates model behavior using consistency score, flip accuracy, and confidence change, providing insights into both correctness and stability. By analyzing model performance under different transformation types, the framework identifies strengths and weaknesses in model robustness.

The contributions of this paper are as follows:
\begin{itemize}
\item A practical batch-based metamorphic testing framework for sentiment analysis models
\item A systematic evaluation of model robustness under multiple linguistic transformations
\item An analysis of model behavior highlighting limitations in handling noise and logical constructs such as negation
\end{itemize}

Overall, this work demonstrates how metamorphic testing can be used as an effective and scalable approach for evaluating robustness in natural language processing systems.

\section{Related Work}

Machine learning systems are traditionally evaluated using statistical metrics such as accuracy, precision, recall, and cross-validation. While these metrics provide a measure of overall model performance, they do not capture how models behave under variations in input data \cite{zhang2019}. In particular, they fail to identify whether a model is robust to small perturbations or transformations, which is critical in real-world applications where input data is often noisy or dynamic.

Metamorphic testing (MT) has been proposed as an effective approach to address the oracle problem in software testing. The core idea of MT is to define metamorphic relations (MRs) that specify how the output should change when the input is transformed. Chen et al. \cite{chen1998} first introduced this concept, demonstrating its effectiveness in testing programs without requiring explicit expected outputs. This idea has since been extended to machine learning systems.

Ding et al. \cite{ding2017} apply metamorphic testing to deep learning frameworks and highlight that traditional validation techniques such as N-fold cross-validation and confusion matrices are insufficient for detecting issues related to dataset quality, implementation correctness, and execution environments. Their work demonstrates that MT can be applied at multiple levels, including datasets, training processes, and inference pipelines.

The METTLE framework proposed in \cite{xie2020} further extends metamorphic testing to machine learning systems, particularly unsupervised learning algorithms. It evaluates model behavior under transformations such as scaling, noise addition, and feature perturbation. However, this approach primarily focuses on structured data and does not directly address challenges specific to natural language processing.

In addition to metamorphic testing, several approaches have been proposed to improve testing coverage for deep neural networks. DeepGauge framework proposed in \cite{ma2018} introduces multi-granularity coverage criteria, including neuron coverage and layer-level coverage. Similarly, surprise adequacy \cite{kim2019} measures how different a test input is from the training data to identify inputs that may expose model weaknesses. While these approaches improve test diversity, they do not explicitly define expected relationships between inputs and outputs.

In the domain of natural language processing, recent work has focused on evaluating robustness through controlled input transformations. The CheckList framework proposed in \cite{ribeiro2020} provides a systematic approach to behavioral testing of NLP models using test cases such as synonym replacement, negation, and paraphrasing. Belinkov and Bisk \cite{belinkov2018} show that neural models are highly sensitive to noise, including typographical errors and character-level perturbations.

Jia and Liang \cite{jia2017} further demonstrate that adversarial examples created through small input modifications can expose significant weaknesses in reading comprehension systems. These findings highlight the importance of testing NLP models under realistic input variations.

Despite these advances, there is still a gap in applying metamorphic testing to NLP using simple, task-specific transformations, consistent evaluation metrics, and scalable batch-based workflows. This paper addresses this by proposing a lightweight metamorphic testing framework for sentiment analysis models.

\section{Problem Statement}

Machine learning models, particularly in natural language processing tasks such as sentiment analysis, are typically evaluated using static datasets and aggregate performance metrics such as accuracy, precision, and F1-score. While these metrics provide a general indication of model performance, they do not capture how the model behaves under realistic variations in input data. In practical applications, input text often contains paraphrasing, typographical errors, informal language, and structural variations, all of which can significantly affect model predictions.

A key limitation in testing machine learning systems is the absence of a reliable test oracle. In many NLP tasks, determining the correct output for every possible input is difficult or subjective. For example, sentiment classification may depend on subtle contextual cues, and different interpretations may be valid. This makes it challenging to directly verify whether a model’s prediction is correct for a given input. As a result, traditional testing approaches are not sufficient for validating model behavior in complex real-world scenarios.

A robust sentiment analysis model should satisfy two important properties. First, it should maintain consistent predictions under transformations that preserve the semantic meaning of the input, such as synonym replacement or minor rephrasing. Second, it should appropriately change its predictions when the semantic meaning is altered, such as through negation or sentiment reversal. Evaluating whether these properties hold requires testing methods that go beyond static datasets and consider input transformations explicitly.

Existing approaches partially address this problem but have important limitations. Metamorphic testing frameworks define relationships between inputs and outputs, but they are often applied to structured data or limited experimental settings. Similarly, NLP robustness studies explore the effects of noise and adversarial inputs, but they are not always grounded in a formal testing framework that defines expected behavior. As a result, there is a lack of systematic approaches that combine the theoretical strengths of metamorphic testing with practical evaluation techniques for NLP models.

Another challenge is scalability. Many existing testing approaches evaluate inputs individually, making them difficult to apply in real-world machine learning pipelines where models process large volumes of data. There is a need for testing methods that can operate efficiently at scale while maintaining consistency in evaluation.

This paper addresses these challenges by proposing a batch-based metamorphic testing framework for sentiment analysis models. The approach systematically applies multiple linguistic transformations to input data and evaluates model behavior using clearly defined metrics. By combining metamorphic testing principles with batch processing and NLP-specific transformations, the proposed method provides a practical and scalable solution for evaluating model robustness under realistic conditions.

\section{Proposed Method}

This paper proposes a lightweight and practical metamorphic testing framework for evaluating the robustness of sentiment analysis models. The approach is designed to systematically assess model behavior under realistic input variations by applying a set of predefined linguistic transformations. Unlike traditional evaluation methods that rely on static datasets, the proposed method focuses on analyzing how predictions change when inputs are modified in controlled ways.

A key design goal of the framework is to balance simplicity, interpretability, and scalability. The method uses simple transformation rules that are easy to implement and understand, while also supporting batch-based evaluation to enable efficient testing of multiple inputs.

\subsection{Overview}

The proposed framework consists of three main stages:

\begin{enumerate}
    \item Selection of input text samples from a labeled sentiment dataset
    \item Generation of transformed inputs using predefined linguistic transformations
    \item Evaluation of model predictions on original and transformed inputs using consistency and robustness metrics
\end{enumerate}

The overall objective is to identify violations of expected behavior, where the model produces inconsistent or incorrect predictions under transformations.

\subsection{Metamorphic Relations}

The core of the proposed method is the definition of metamorphic relations (MRs), which specify how model outputs are expected to change when inputs are transformed. These relations are categorized into two types based on the nature of the transformation.

\subsubsection{Invariance Relations}

Invariance relations correspond to transformations that preserve the semantic meaning of the input. Under such transformations, the model is expected to produce the same prediction.

The following transformations are used:

\begin{itemize}
    \item \textbf{Synonym Replacement:} Words in the input text are replaced with semantically equivalent synonyms. For example, the sentence ``The movie was good'' may be transformed to ``The movie was great.'' Since the meaning remains unchanged, the predicted sentiment should also remain the same.

    \item \textbf{Typo Injection:} Minor typographical errors are introduced into the text, such as character deletion or substitution (e.g., ``service'' $\rightarrow$ ``servce''). These transformations simulate real-world user input, where spelling mistakes are common.

    \item \textbf{Word-Order Variation:} The order of words in the sentence is modified without changing the overall meaning (e.g., ``I really liked this movie'' $\rightarrow$ ``This movie I really liked''). This tests whether the model captures contextual relationships rather than relying strictly on word position.
\end{itemize}

These transformations are chosen because they represent common variations in natural language that should not alter the underlying sentiment.

\subsubsection{Directional Relations}

Directional relations correspond to transformations that intentionally change the semantic meaning of the input. Under such transformations, the model is expected to change its prediction accordingly.

\begin{itemize}
    \item \textbf{Negation Transformation:} A negation term is introduced to reverse the sentiment of the sentence (e.g., ``The movie was good'' $\rightarrow$ ``The movie was not good''). This transformation tests whether the model correctly interprets logical modifiers that alter meaning.
\end{itemize}

Negation is particularly important because it requires compositional understanding of language, which is a known challenge for many NLP models.

\subsection{Batch-Based Evaluation}

To improve scalability and efficiency, the proposed method evaluates inputs in batches. Given a set of input texts, transformations are applied to generate corresponding modified inputs. The model is then evaluated on both the original and transformed inputs in a batched manner.

Batch processing offers several advantages:
\begin{itemize}
    \item improved computational efficiency compared to evaluating inputs individually
    \item consistent evaluation across multiple inputs and transformations
    \item compatibility with real-world inference pipelines, where models process data in batches
\end{itemize}

This design choice makes the framework practical for large-scale testing scenarios.

\subsection{Evaluation Metrics}

The framework uses three primary metrics to evaluate model behavior:

\begin{itemize}
    \item \textbf{Consistency Score:} Measures the percentage of inputs for which the model prediction remains unchanged under invariance transformations. A high consistency score indicates robustness to semantic-preserving variations.

    \item \textbf{Flip Accuracy:} Measures the percentage of inputs for which the model correctly changes its prediction under directional transformations such as negation. This metric evaluates the model’s ability to capture changes in semantic meaning.

    \item \textbf{Confidence Change:} Measures the difference in model confidence between original and transformed inputs. This metric provides insight into the stability of model predictions, even when the predicted label remains unchanged.
\end{itemize}

Together, these metrics provide a comprehensive view of model robustness, capturing both correctness and stability.

The proposed method provides a simple yet effective framework for evaluating robustness in sentiment analysis models. By combining metamorphic testing principles with NLP-specific transformations and batch-based evaluation, the approach enables systematic analysis of model behavior under realistic input variations. The framework is easy to implement, interpretable, and scalable, making it suitable for practical deployment in machine learning testing workflows.

\section{Experimental Setup}

This section describes the dataset, model, implementation details, and evaluation procedure used to assess the proposed metamorphic testing framework.

\subsection{Dataset}

To evaluate the robustness of the sentiment analysis model, we use a subset of a publicly available sentiment classification dataset consisting of short text reviews labeled as positive or negative. The dataset contains natural language sentences with varying levels of complexity, making it suitable for evaluating model behavior under linguistic transformations.

For this study, a subset of approximately 200 samples is selected. The goal is not to maximize predictive performance, but to analyze model behavior under controlled input variations. Using a smaller dataset allows for detailed inspection of individual cases and facilitates qualitative analysis through case studies.

The selected samples include a mix of clearly positive, clearly negative, and moderately ambiguous sentences. This diversity ensures that the evaluation captures both easy and challenging scenarios for the model.

\subsection{Model}

We use a pretrained transformer-based sentiment classification model based on DistilBERT~\cite{sanh2019distilbert}. DistilBERT is a lightweight version of BERT that retains most of its performance while reducing computational complexity. It uses self-attention mechanisms to capture contextual relationships between words in a sentence.

The model has been fine-tuned on the SST-2 dataset, which is widely used for sentiment analysis tasks. This ensures that the model has learned general sentiment patterns and can serve as a representative example of modern NLP models.

The model takes raw text as input and produces:
\begin{itemize}
    \item a predicted sentiment label (positive or negative)
    \item a confidence score representing the model's certainty
\end{itemize}

Using a pretrained model reflects real-world deployment scenarios, where models are often reused rather than trained from scratch.

\subsection{Implementation Details}

The experiment is implemented in Python using the Hugging Face Transformers library. The pipeline API is used for efficient model inference.

Text transformations are implemented using simple rule-based methods:
\begin{itemize}
    \item synonym replacement using predefined word mappings
    \item typo injection through character-level modifications
    \item word-order variation using basic reordering rules
    \item negation insertion using template-based sentence modification
\end{itemize}

All inputs are processed in batches to improve efficiency and ensure consistency across evaluations. Batch processing aligns with typical machine learning inference workflows and enables scalable testing.

\subsection{Testing Procedure}

The evaluation follows a structured metamorphic testing workflow:

\begin{enumerate}
    \item \textbf{Input Selection:} A subset of sentences is selected from the dataset.
    
    \item \textbf{Transformation Generation:} For each input sentence, multiple transformed versions are generated using predefined transformations.
    
    \item \textbf{Model Evaluation:} The model is applied to both original and transformed inputs to obtain predictions and confidence scores.
    
    \item \textbf{Comparison:} Predictions are compared to identify whether expected metamorphic relations are satisfied or violated.
    
    \item \textbf{Metric Computation:} Metrics such as consistency score, flip accuracy, and confidence change are computed for each transformation type.
\end{enumerate}

\subsection{Design Considerations}

The experimental design prioritizes simplicity and interpretability. The transformations are intentionally lightweight and easy to understand, allowing clear interpretation of model behavior.

Another important consideration is realism. The chosen transformations reflect common variations in real-world text, such as spelling errors, paraphrasing, and negation. This ensures that the evaluation results are meaningful for practical applications.

Finally, the use of batch-based evaluation improves scalability and makes the framework suitable for integration into real-world testing pipelines.

\section{Experimental Results}

This section presents the results of applying the proposed metamorphic testing framework to the sentiment analysis model. The evaluation focuses on measuring model robustness under different linguistic transformations using consistency score, flip accuracy, and confidence change.

\begin{table}[h]
\centering
\caption{Model Performance Across Transformations}
\begin{tabular}{|l|l|c|}
\hline
\textbf{Transformation} & \textbf{Metric} & \textbf{Score (\%)} \\
\hline
Synonym Replacement & Consistency Score & 100.0 \\
Typo Injection & Consistency Score & 98.5 \\
Word Order Change & Consistency Score & 96.5 \\
Negation & Flip Accuracy & 28.5 \\
\hline
\end{tabular}
\label{tab:main_results}
\end{table}

Table~\ref{tab:main_results} summarizes the model performance across different transformations. The model achieves high consistency for synonym replacement and word-order variation, indicating robustness to semantic-preserving changes. However, performance drops significantly for negation, highlighting difficulty in handling logical transformations.

\begin{table}[h]
\centering
\caption{Average Confidence Change Across Transformations}
\begin{tabular}{|l|c|}
\hline
\textbf{Transformation} & \textbf{Avg Confidence Change} \\
\hline
Synonym Replacement & 0.0011 \\
Typo Injection & 0.0045 \\
Word Order Change & 0.0005 \\
Negation & 0.0145 \\
\hline
\end{tabular}
\label{tab:confidence}
\end{table}

Table~\ref{tab:confidence} shows the average change in confidence across transformations. Even when predictions remain unchanged, confidence often decreases, indicating reduced certainty under transformed inputs.

\begin{figure}[h]
\centering
\includegraphics[width=\linewidth]{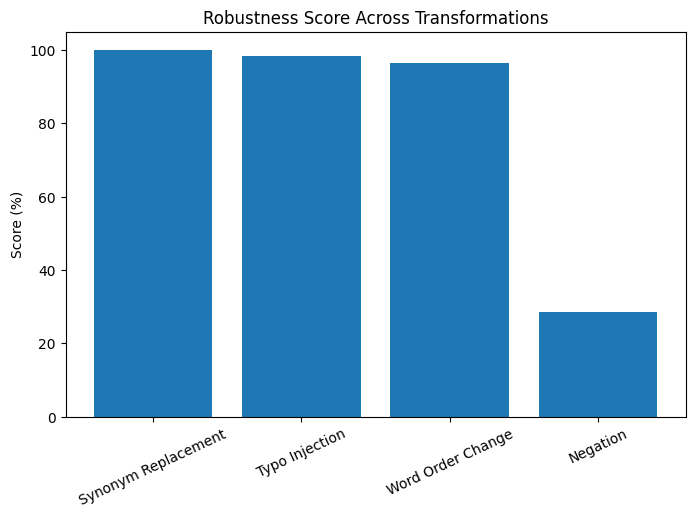}
\caption{Consistency score across different transformations}
\label{fig:robustness}
\end{figure}
Figure~\ref{fig:robustness} illustrates the consistency scores across transformations. The model maintains high consistency for semantic-preserving transformations but shows reduced robustness for transformations involving noise and logical changes.

\begin{figure}[h]
\centering
\includegraphics[width=\linewidth]{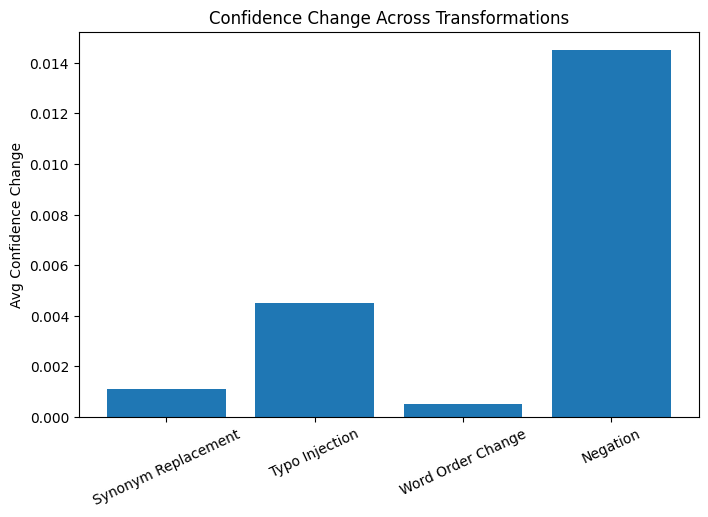}
\caption{Confidence variation across transformations}
\label{fig:confidence}
\end{figure}
Figure~\ref{fig:confidence} shows how model confidence varies across transformations. The increase in confidence fluctuation for negation and typo transformations indicates instability in model predictions.

\begin{figure}[h]
\centering
\includegraphics[width=\linewidth]{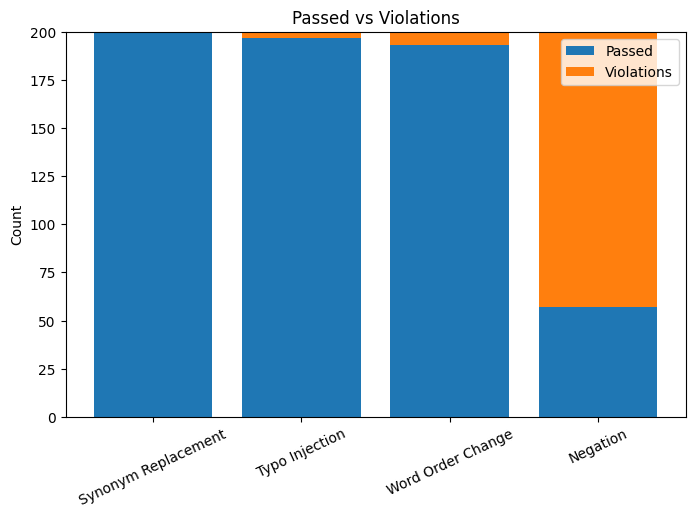}
\caption{Number of test cases passed versus violated}
\label{fig:violations}
\end{figure}

Figure~\ref{fig:violations} highlights that most violations occur under negation, indicating failure to capture logical transformations.

\begin{table*}[t]
\centering
\caption{Sample Metamorphic Violations}

\begin{tabular}{p{2cm} p{4.2cm} p{4.2cm} c c }
\hline
\hline
\textbf{Transformation} & \textbf{Original Text} & \textbf{Transformed Text} & \textbf{Orig. Label} & \textbf{Trans. Label} \\
\hline
\hline
Negation & it 's a bad thing when a movie has about as much substance as its end credits blooper reel . &
it 's a not bad thing when a movie has about as much substance as its end credits blooper reel . &
NEG & POS \\
\hline

Negation & the heavy-handed film is almost laughable as a consequence . &
not the heavy-handed film is almost laughable as a consequence . &
NEG & POS \\
\hline

Typo & pumpkin means to be an outrageous dark satire on fraternity life , but its ambitions far exceed the abilities ... &
pumpkin means to be an outrageous dark satire on fraternity life , but its ambitions far exced the abilities ... &
POS & NEG \\
\hline

Word Order & it seems like i have been waiting my whole life for this movie and now i ca n't wait for the sequel . &
seems it like i have been waiting my whole life for this movie and now i ca n't wait for the sequel . &
POS & NEG \\
\hline

Word Order & the script kicks in , and mr. hartley 's distended pace and foot-dragging rhythms follow . &
script the kicks in , and mr. hartley 's distended pace and foot-dragging rhythms follow . &
POS & NEG \\
\hline

Typo & though perry and hurley make inspiring efforts to breathe life into the disjointed , haphazard script ... &
though perry and hurley make inspiring efforts to brethe life into the disjointed , haphazard script ... &
POS & NEG \\
\hline

\end{tabular}
\label{tab:qualitative_examples}
\end{table*}

Table~\ref{tab:qualitative_examples} presents representative examples where small input transformations lead to incorrect sentiment predictions. These cases highlight the model's sensitivity to typographical noise, word-order variations, and especially negation, where the intended semantic meaning is not correctly captured.

\subsection{Performance Across Transformations}

The results show that the model exhibits varying levels of robustness depending on the type of transformation applied.

For synonym replacement, the model achieves very high consistency, indicating that it is able to preserve predictions when semantically equivalent words are substituted. This suggests that the model has learned meaningful semantic representations and can generalize across similar vocabulary.

Word-order variation also results in high consistency scores in most cases. This indicates that the model is capable of capturing contextual relationships beyond strict word positioning, likely due to the use of attention mechanisms in transformer architectures.

In contrast, typo injection leads to a noticeable decrease in consistency. Even minor character-level changes can result in incorrect predictions or reduced confidence. This highlights the model's sensitivity to surface-level noise and suggests limitations in handling real-world text inputs that may contain spelling errors or informal language.

\subsection{Negation Handling}

Negation presents a significant challenge for the model. The results show that the flip accuracy for negation transformations is relatively low, indicating that the model often fails to correctly reverse sentiment when negation is introduced.

This behavior suggests that the model relies heavily on sentiment-bearing words and may not fully capture the compositional structure of language. As a result, phrases such as ``not good'' may still be interpreted as positive due to the presence of strong positive words.

\subsection{Confidence Analysis}

In addition to prediction accuracy, confidence scores provide insights into the stability of the model.

Across all transformation types, a general trend is observed where confidence decreases when inputs are modified, even if the predicted label remains unchanged. This indicates that the model becomes less certain under transformed inputs, reflecting sensitivity to input variations.

In some cases, incorrect predictions are made with relatively high confidence, highlighting overconfidence in the model. This behavior can be problematic in real-world applications where confidence is used to assess prediction reliability.

\subsection{Summary of Results}

Overall, the results demonstrate that the model performs well under semantic-preserving transformations such as synonym replacement and word-order variation, but struggles with noisy inputs and logical transformations such as negation.

These findings emphasize that high accuracy on static datasets does not necessarily imply robustness under realistic conditions. The proposed metamorphic testing approach provides a more comprehensive evaluation by revealing how model behavior changes under different types of input transformations.

\section{Analysis and Discussion}

The results show that the sentiment analysis model exhibits varying levels of robustness depending on the type of transformation applied.

Synonym replacement achieves perfect consistency, indicating that the model can effectively preserve predictions when semantically equivalent words are substituted. Typo injection introduces minor failures, suggesting sensitivity to character-level noise. Word-order variation slightly reduces consistency, indicating partial dependence on sentence structure and positional information.

The most significant limitation is observed in negation handling, where the model frequently fails to correctly reverse sentiment. This suggests that the model relies heavily on sentiment-bearing words while struggling to capture compositional semantics and logical modifiers such as ``not.''

Confidence analysis further highlights model instability. Even when predictions remain unchanged, confidence consistently decreases after transformation. This indicates that the model becomes less certain when inputs are modified. In several cases, incorrect predictions are produced with relatively high confidence, revealing overconfidence in model behavior.

These observations can be explained by the underlying architecture of transformer-based models. Such models rely on tokenization and contextual embeddings to represent text. While this enables strong semantic understanding, it also creates sensitivity to surface-level perturbations such as typographical errors, which may disrupt token representations.

The difficulty in handling negation reflects a broader challenge in natural language processing. Negation requires compositional understanding, where sentence meaning depends on interactions between words rather than individual tokens alone. Models may therefore overemphasize strongly positive or negative words while underweighting negation terms.

From a practical perspective, these findings highlight the importance of robustness testing beyond traditional accuracy metrics. In real-world applications such as customer feedback analysis or content moderation, input text frequently contains noise, informal language, and structural variations. A model that is sensitive to such changes may produce unreliable predictions despite achieving high benchmark accuracy.

Confidence-based analysis also provides additional insights into prediction reliability. Confidence scores can help identify uncertain predictions and may support future approaches involving uncertainty estimation or confidence-aware filtering mechanisms.

\section{Case Study: Analysis of Model Failures}

To further understand the behavior of the sentiment analysis model, we examine specific examples where the model fails under different transformations. These case studies provide qualitative insights that complement the quantitative results presented earlier.

\subsection{Failure under Typo Injection}

Consider the original sentence:

\begin{quote}
``The movie was absolutely wonderful and engaging.''
\end{quote}

The model correctly predicts a positive sentiment with high confidence. However, after introducing minor typographical errors:

\begin{quote}
``The movie was absolutly wonderfull and engaging.''
\end{quote}

the model prediction changes to negative in some cases or shows a significant drop in confidence.

This behavior suggests that the model is sensitive to character-level perturbations. Transformer-based models rely on tokenization to convert text into subword units. Typographical errors can disrupt this process, leading to incorrect token representations and degraded performance. This highlights a limitation in handling noisy real-world inputs.

\subsection{Failure under Word-Order Variation}

Consider the sentence:

\begin{quote}
``I really enjoyed this movie and would recommend it.''
\end{quote}

After applying word-order variation:

\begin{quote}
``This movie I really enjoyed and would recommend it.''
\end{quote}

the model generally maintains the correct prediction. However, in more complex sentences with multiple clauses, reordering can occasionally lead to incorrect predictions.

This suggests that while the model captures contextual relationships, it may still rely partially on positional patterns learned during training. Changes in sentence structure can alter attention patterns, leading to inconsistencies in prediction.

\subsection{Failure under Negation}

Negation presents the most significant challenge. Consider the sentence:

\begin{quote}
``The movie was good and entertaining.''
\end{quote}

After applying negation:

\begin{quote}
``The movie was not good and entertaining.''
\end{quote}

the model often fails to correctly reverse the sentiment and continues to predict a positive label.

This behavior indicates that the model does not fully capture compositional semantics. Instead, it appears to rely heavily on sentiment-bearing words such as ``good'' and ``entertaining,'' without adequately accounting for negation terms. This limitation has been observed in prior NLP studies and remains an open challenge in language understanding.

\subsection{Insights from Case Study}

These examples highlight key limitations of the model:

\begin{itemize}
    \item Sensitivity to surface-level noise due to reliance on tokenization
    \item Partial dependence on word order and positional patterns
    \item Difficulty in handling compositional semantics, particularly negation
\end{itemize}

The case study reinforces the importance of robustness testing beyond aggregate metrics. While overall accuracy may appear high, these examples demonstrate that the model can fail under realistic input variations. Metamorphic testing provides a structured way to uncover such failures and analyze their underlying causes.

\section{Conclusion and Future Work}

This paper presented a lightweight and practical framework for evaluating the robustness of sentiment analysis models using metamorphic testing. Unlike traditional evaluation methods that rely on static datasets and aggregate metrics, the proposed approach focuses on analyzing model behavior under controlled input transformations. By applying batch-based metamorphic testing with multiple linguistic transformations, the framework provides a more comprehensive understanding of how models behave under realistic conditions.

The experimental results demonstrate that the sentiment analysis model exhibits varying levels of robustness depending on the type of transformation applied. The model performs strongly under semantic-preserving transformations such as synonym replacement and word-order variation, indicating its ability to capture contextual meaning. However, the model is sensitive to typographical errors, highlighting limitations in handling noisy input. More importantly, the model struggles to correctly handle negation, revealing weaknesses in capturing compositional semantics and logical structure in language. The analysis and case study further show that confidence scores provide valuable insights into model stability. Even when predictions remain unchanged, reduced confidence under transformed inputs indicates underlying uncertainty. Additionally, incorrect predictions with high confidence reveal overconfidence issues, which can be problematic in real-world applications.


\noindent \textit{Future Work --} There are several directions for future work. The current study uses manually defined transformations. Future work can explore automated generation of metamorphic relations using techniques such as paraphrase generation or language models, enabling more diverse and scalable testing. Additional linguistic transformations such as sarcasm, domain-specific language, and multi-sentence context can be incorporated to better reflect real-world usage scenarios. Finally, the proposed framework can be integrated into continuous testing pipelines for machine learning systems. This would enable ongoing monitoring of model robustness during deployment and help ensure reliability as models are updated or exposed to new data.

\bibliographystyle{ieeetr}
\bibliography{ref}

\end{document}